\documentclass[journal]{IEEEtai}

\usepackage[numbers]{natbib}

\usepackage[colorlinks,urlcolor=blue,linkcolor=blue,citecolor=blue]{hyperref}

\usepackage{color,array, soul}

\usepackage{graphicx}

\usepackage{caption}
\usepackage{textcomp}
\usepackage{amsmath,amssymb,amsfonts}
\usepackage{algorithmic}
\usepackage[ruled,vlined]{algorithm2e}

\PassOptionsToPackage{hyphens}{url}

\newlength\myindent
\newcommand\bindent{%
  \begingroup
  \setlength{\itemindent}{\myindent}
  \addtolength{\algorithmicindent}{\myindent}
}
\newcommand\eindent{\endgroup}

\begin{document}

\title{Quantifying Explainability of Saliency Methods in Deep Neural Networks with a Synthetic Dataset}

\author{Erico Tjoa, Guan Cuntai, \IEEEmembership{Fellow, IEEE}
\thanks{This work is done under the Alibaba-NTU Talent Program.}
\thanks{Erico is affiliated to Alibaba HealthTech division and Nanyang Technological University at the time of writing. }
\thanks{Guan Cuntai is a professor at Nanyang Technological University, School of Computer Science and Engineering. }
}

\maketitle

\begin{abstract}
Post-hoc analysis is a popular category in eXplainable artificial intelligence (XAI) study. In particular, methods that generate heatmaps have been used to explain the deep neural network (DNN), a black-box model. Heatmaps can be appealing due to the intuitive and visual ways to understand them but assessing their qualities might not be straightforward. Different ways to assess heatmaps' quality have their own merits and shortcomings. This paper introduces a synthetic dataset that can be generated adhoc along with the ground-truth heatmaps for more objective quantitative assessment. Each sample data is an image of a cell with easily recognized features that are distinguished from localization ground-truth mask, hence facilitating a more transparent assessment of different XAI methods. Comparison and recommendations are made, shortcomings are clarified along with suggestions for future research directions to handle the finer details of select post-hoc analysis methods. Furthermore, mabCAM is introduced as the heatmap generation method compatible with our ground-truth heatmaps. The framework is easily generalizable and uses only standard deep learning components.
\end{abstract}

\begin{IEEEImpStatement}
We provide a synthetic dataset with clear, unambiguous features. We demonstrate that heatmap-based XAI methods are not necessarily producing heatmaps that correspond to human intuition of ``explanation". 
\end{IEEEImpStatement}


\section{Introduction}

EXplainable artificial intelligence (XAI) has been gathering attention in the artificial intelligence (AI) and machine learning (ML) community recently. The trend was propelled by the success of deep neural network (DNN), especially convolutional neural network (CNN) in image processing. DNN has been considered a blackbox because the mechanism underlying its remarkable performance is not well understood. XAI research has thus developed in many different directions. Among them is the saliency method, where heatmaps are generated to point at where an AI model is ``looking at" as it makes a decision or prediction. The heatmaps have been treated as explanations, and they are desirable because are compatible with human's visual comprehension, easy to read and interpret. 

In the most ideal scenario, ground-truth heatmaps are available, and they provide us with clues on how to fix an algorithm that produces wrong predictions. The sub-optimal heatmaps facilitate an analysis that reveals the problematic parts of the algorithm. Once these faulty parts are identified, the algorithm can be iteratively improved, and then new heatmaps converge closer towards the ground-truth heatmaps. However, ground-truth heatmaps are sometimes not available. Furthermore, many of the formulas used to generate heatmaps are given using heuristics, e.g. the baseline obtained by the gradient method \cite{Simonyan14a} or by setting negative signals/weights to zeros \cite{PixelWiseLRP}, hence their relations to the groundtruth might not be clear.

Various XAI methods have been designed and evaluated in many different ways. However, evaluation methods do not necessarily reflect the relevant need for explanations, especially because the correctness of explanations is context dependent and can be subjective. In this paper, our goal is \textbf{to measure the correctness of explanation} as objectively as possible in the context of image classification. This is done by using a synthetic dataset with in-built ground-truth heatmaps whose interpretability matches human intuition. The novelty lies in the generation of unambiguous discrete-valued attribution heatmaps as explanation and the introduction of a stringent performance metric that counts ``hits" and ``misses" without ambiguity. The dataset is also customizable and augmented with noisy variations of basic shapes achievable by closed-form equations (such as circles, rectangles etc). Besides, ground-truth heatmaps are automatically generated alongside the image data and labels, avoiding the laborious process of manually marking heatmap features. Finally, we will introduce mabCAM, an easily generalizable heatmap generation method that is compatible the aforesaid dataset.
 
In section \ref{sect:related}, we review how XAI methods have been evaluated in the literature and focus on the challenges in their interpretation. In section \ref{subsection:data}, we introduce the aforementioned synthetic dataset. For this paper, specifically a 10-class dataset is used to compare 9 different XAI methods. Each data sample consists of an object with a simple shape and its corresponding heatmap designed to be numerically unambiguous. The shapes and colours are computed through Algorithm 1 and 2 using modifications of standard mathematical formulas (equation of ellipse etc); github link is available for details. More precisely, the correctness of heatmaps can be verified in a simple and objective way through the computation of recall and precision. The rest of section \ref{section:method} describes the implementation of neural network training for ResNet34 (henceforth ResNet), AlexNet and VGG, followed by validation and evaluation processes, followed by the description of \textit{five-band-score}, a metric defined to capture quantities such as recall and precision that take into account the distinct meaningful regions in heatmaps. We have also introduced \textit{mabCAM}, a generic framework to generate heatmaps akin to semantic segmentation process that is compatible with the framework described by our synthetic heatmaps. Section \ref{section:discussion} discusses the recall-precision results and ROC curves. Finally, we conclude with recommendations and provide some caveats. Note: though heatmaps are sometimes interchangeably called saliency map, we only refer to them as heatmaps here because we want to distinguish them from XAI method whose name is Saliency. 

\begin{figure*}[h]
\centerline{\includegraphics[width=0.9\textwidth]{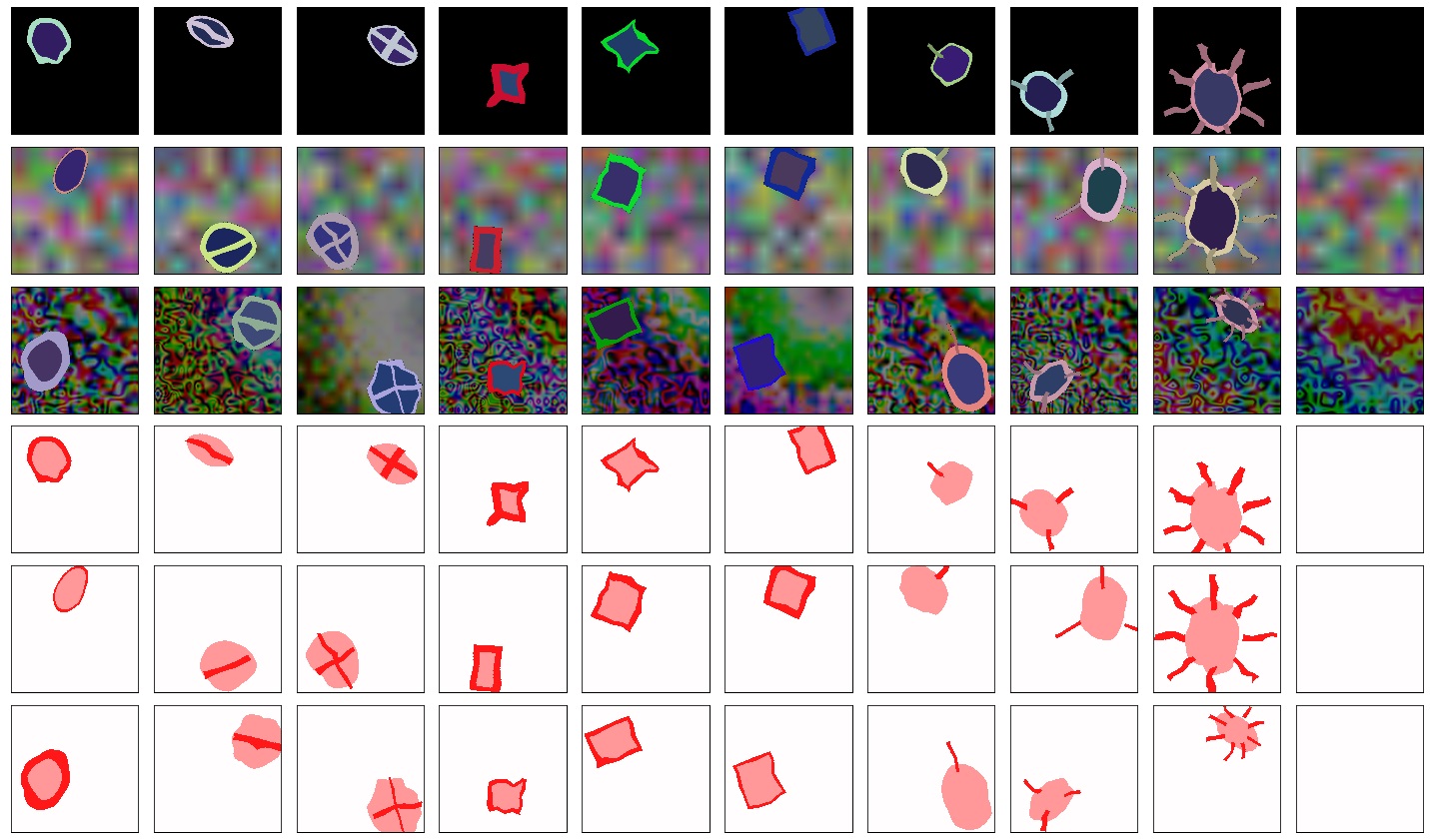}}
\caption{The first row shows 10 different types of shapes that can be generated by our algorithm, placed in background type 1, i.e. dark background. We refer to the different types of cells as cells type 0 to 8, where 0,1,2 are circular, 3,4,5 are rectangular and 6,7,8 are circular with one, three and eight tails respectively. Their alternative names in the codes are CCell (0), CCellM (1), CCellP (2), RCell (3), RCellB (4), RCellC (5), CCellT (6), CCellT3 (7), CCellT8 (8) respectively. C denotes circular cell, R rectangular, T tails, M minus, P plus. The last column (or type 9) does not contain any cell. The second and third rows are similar to the first row, except they are placed on background type 2 and 3 respectively. Row 4, 5, 6 are the ground-truth heatmaps for row 1, 2, 3 respectively. The region colored light-red corresponds to localization information, while dark red region corresponds to distinguishing features. For example, column 1 and 2 can be distinguished by the presence of the bar across the circular cell. Columns 4, 5, 6 differ only in their dominant colors of their rectangular borders, and thus the distinguishing features are their borders.}
\label{fig:maingallery}
\end{figure*}

\begin{figure*}
\centerline{\includegraphics[width=\textwidth]{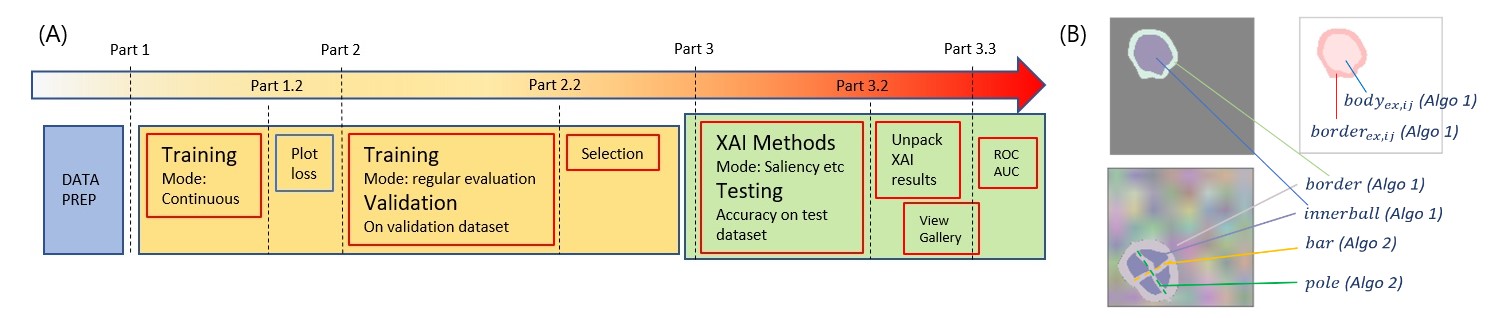}}
\caption{(A) Workflow illustrating the process starting from data generation to the generation of heatmaps gallery. (B) Illustration showing how parts of the cells are generated in algorithms 1 and 2. }
\label{fig:workflow}
\end{figure*}

\section{Related Works}
\label{sect:related}
In this section, we provide a survey of several XAI papers and briefly discuss how their results are quantified. 

\subsection{Regarding Heatmap Ground-truths}
Our idea of using synthetic data for evaluating XAI methods resembles the synthetic flower dataset in \cite{oramas2018visual}. The flower dataset comes with ground-truth masks for discriminating features. The paper measures the correctness of heatmap explanations using IoU (Intersection over Union) between thresholded heatmaps and the ground-truths. Our dataset also provides ground-truth masks, but we provide different labels for two separate notions ``discriminating features" and ``localization" (respectively shown as dark red and light red regions in fig. \ref{fig:maingallery}). We assess the quality of heatmaps using precision, recall and ROC instead, though the quantities are slightly modified as we distinguish localization from feature discrimination.

The paper that introduces SmoothGrad \cite{DBLP:journals/corr/SmilkovTKVW17} mentioned that, at the time of its publication, there was no ground-truth to allow for quantitative evaluation of heatmaps.  It then proceeds with 2 qualitative evaluations instead. As of now, even though there are many different datasets available for AI and ML researches, the corresponding ground-truth explanations (such as heatmaps) are typically not available. To address this problem, we directly provide unambiguous ground-truth heatmaps.

\subsection{XAI and their Evaluations}
Heatmap-based XAI methods have been evaluated using several different metrics. Many of them suffer from various problems, especially correctness w.r.t human observation e.g. CHR problem (defined later), and, in an extreme case, \textit{the pointing game}. By contrast, this work provides a very unambiguous way to measure the correctness of heatmaps as XAI \textit{explanation}. This paper also serves as an instance to reiterate that existing XAI methods can be sensitive to datasets, thus cautioning users on their context dependence.

The experimental study \cite{Li2020QuantitativeEO} have explored many existing metrics, through which they compare different heatmap-based XAI methods on several DNN architecture and datasets. The metrics include faithfulness, sensitivity and stability, some with modifications.

Saliency, deconvolution and LRP (Layerwise Relevance Propagation) have been compared \cite{7552539} using the ``most relevant first" (MoRF) heatmap evaluation framework, where the quantity Area under Perturbation Curve (AOPC) is computed. Heatmap pixels are ordered according to importance \(O=(r_1,r_2,\dots)\). The original image is perturbed by replacing the most important pixels starting from \(r_1\), and then its AOPC is computed progressively with more perturbations. There is a \textit{computational vs human relevance} (CHR) problem, i.e. what is computationally most relevant may not correspond to what human finds relevant, especially when \(r_k\) can be a single isolated pixel. The correctness of ordering \(O\) is not addressed.

CAM \cite{CAM} and GradCAM \cite{GradCAM} heatmaps are shown to improve the localization on ILSVRC (ImageNet Large Scale Visual Recognition Challenge) datasets. By observing the change in the log odd scores after deleting image pixels, the relevance of image pixels to the decision or prediction of a model can be determined as well \cite{pmlr-v70-shrikumar17a}. GradCAM paper demonstrates through human studies that its heatmaps help increase human's performance in categorical tasks. CAM paper and many other heatmap-based XAI papers do not report report similar human studies. CHR problem may be present, as only computational relevance is presented. 

The earlier paper on layerwise relevance propagation (LRP) \cite{PixelWiseLRP} introduces the epsilon-LRP. It displays heatmaps generated from many sample data, although many heatmaps do not appear to visually demonstrate good consistency in their pixel-wise assignment of values (different improvements have since been suggested). Tests were conducted on the effect of transformation on the images, for example, by flipping MNIST digits, and \textit{mean prediction} is defined to assess the method after interchanging pixels systematically based on relevance computed by LRP. Still, the paper itself mentions that the analysis is semi-quantitative. Furthermore, quantitative results presented mainly analyzes computational relevance (CHR problem). Other methods that may suffer from CHR problem include ROAR (RemOve And Retrain) \cite{NIPS2019_9167} (consider the correctness of ordering \(\{e_i^o\}^N_{i=1}\)) and SWAG \cite{Hartley_2021_WACV} (see e.g. their black swan saliency).

Two simple sanity checks \cite{NIPS2018_8160} have been proposed for saliency maps (though they are easily applicable for others): (1) test whether algorithm parameters affect the heatmap output (2) test whether data ordering affect the output. The first sanity check is performed by randomizing weights layer by layer and then computing similarity metrics such as structural similarity index (SSIM), histogram of gradients (HOGs) and rank correlations. It reports possibly severe problems with existing methods. For example, after a DNN's weight parameters are randomized in all layers, GradCAM still produces heatmaps with high similarity values when compared to heatmaps produced by the DNN with no or few randomization, as though the DNN itself is irrelevant. Besides, the paper also reports that epsilon-LRP, DeepLift \cite{pmlr-v70-shrikumar17a} and integrated gradients \cite{10.5555/3305890.3306024} return a large part of the input image, possibly confusing inputs with the values relevant to explanations. Using DeepLIFT on our synthetic dataset, on the contrary, show more favorable results, as background noises appear to be filtered away to a good extent. This could indicate sensitivity to the dataset.

Likewise, several heatmap methods are evaluated by measuring similarity w.r.t  \textit{sensitivity-n} property \cite{ancona2018towards}. The sum of attribution is compared to the same sum but with \(n\) features removed. The paper presents several useful empirical observations from their XAI evaluations. We similarly provide more observations on heatmaps as explanations at the end of this paper. Similarity metrics are also used with the above-mentioned sanity checks in \cite{Sixt2020WhenEL}.

The \textit{pointing game} has been used for evaluating the performance of localization algorithm. In XAI evaluations, they are used in \cite{8237633,9157775}. Looking more closely, this metric is ``extremely loose" because it considers a hit whenever the maximum value lies inside the desired ground-truth mask region (even a single pixel counts). Using a single pixel to evaluate the quality of explanations may be  questionable, unless the mask area is always significantly small compared to the image size. By contrast, we measure recall and precision over a large regions of pixels in the images.

Another experimental study on the quantitative evaluations of heatmap-based methods \cite{10.1145/3447548.3467148} listed several issues in existing methods as well (e.g. faithfulness has out-of-distribution issues). It introduces IntIoSR, again another metric to compute overlap with ground-truth heatmaps; our paper again differs from such ``localization only" approach, since we distinguish localization from discriminative features. Likewise, \cite{ARRAS202214} performs such localization measurement using Relevance Mass Accuracy, though it also includes Relevance Rank Accuracy. Our metric is similar in the sense that we compute pixelwise hit/miss.

\subsection{Other XAIs and Applications}
XAI methods that are not focused on generating heatmaps have also been developed. This paper is mainly concerned with quantitative heatmaps comparison, but we may still benefit from different types of evaluations of XAI performance. Local interpretable model-agnostic explanation (LIME) \cite{10.1145/2939672.2939778} is introduced to find a locally faithful interpretable model that represents well the model under inspection, regardless of the latter's architecture (i.e. is agnostic). By comparing LIME with obviously interpretable models such as decision trees and sparse logistic regression, in particular using recall value, the quality of feature importance obtained using LIME can be assessed. Experiments on Concept Activation Vectors (section 4.3 of \cite{conf_icml_KimWGCWVS18}) include quantitative comparison of the information used by a model when a ground-truth caption is embedded into the image. In some cases, the caption is used by the model for decision-making, but in other cases, only the image concept is used. Furthermore, human-subject experiments are also conducted to test the importance of the saliency mask, showing that heatmaps only marginally help human make decision. It also shows that heatmaps can even be misleading. There has also been other similar sentiments, for example in the caption of fig. 2 in \cite{Rudin2019}, where the utility of a heatmap is called into question. Network dissection frameworks \cite{8099837,8417924,Bau_2020} have been used to evaluate DNN by counting the number of \textit{unique detectors} and measuring the IoU between feature maps and segmentation mask. There are other methods not covered here, and we refer the readers to survey papers like \cite{BARREDOARRIETA202082,8631448,9233366}.

Applications of XAI methods have emerged in several fields e.g. in network traffic analysis \cite{app11104697,9490313} and the medical field, where evaluation of heatmaps has been performed in different ways. A study on MRI-based Alzheimer's disease classification \cite{10.1007/978-3-030-33850-3_1} computes the L2 norm between average heatmaps generated by different XAI methods and compares the performance of three other different metrics. Ground-truth heatmaps are sometimes available, for example in the diagnosis of lung nodules \cite{10.1007/978-3-030-33850-3_5} where recall values can be directly computed between the reference features (ground-truth) and the heatmaps generated by different XAI methods. Different kinds of ground-truth have been obtained using specialized method, such as NeuroSynth in \cite{DBLP:journals/corr/abs-1810-09945} for analyzing neuroimaging data. Some parts of the evaluation are qualitative (such as group-level evaluation), though the paper uses F1-score to evaluate the heatmap, thus naturally including recall and precision concepts in the evaluation. Other applications of XAI methods, especially heatmaps, in the medical field are for example \cite{DBLP:journals/corr/abs-1901-07031, 10.1007/978-3-030-00928-1_55, 10.1007/978-3-030-00931-1_24, 10.1007/978-3-030-00928-1_56, 10.1007/978-3-030-00934-2_29,  Tang2019NatureAlz,10.1117/12.2549298,DBLP:journals/corr/abs-1805-08403,10.1007/978-3-030-33850-3_3}.

\section{Data and Methodology}
\label{section:method}
This section describes the workflow starting from data generation, network training, network performance evaluation, heatmap generation and evaluating generated heatmaps with common quantities. The workflow is shown in fig. \ref{fig:workflow}(A), closely following the sequence of commands run in the package of python codes \footnote{\url{https://github.com/etjoa003/explainable_ai/tree/master/xai_basic}} provided. Some details, such as the algorithms needed to generate each sample data, can be traced from the tutorials available as jupyter notebook included in the package of codes. 

\subsection{Dataset}
\label{subsection:data}

\begin{algorithm}
\begin{algorithmic}
\caption{build\_basic\_ball\_body($x_0,y_0,r,t$ etc) for type 0 cell, see fig. \ref{fig:maingallery}. *More descriptions in appendix.}\label{buildbasicball}
\STATE $x,y \gets mesh grid$ 
\STATE $d\gets \sqrt{x^2+(y/y_{s})^2} + noise$
\STATE $border\gets (d\leq r)*(d\geq r-t)$
\STATE $innerball\gets (d < r-t)$
\STATE rotate by \(\theta\)
\STATE shift center to $(x_0,y_0)$
\STATE $ball\gets border + innerball$
\STATE make\_explanation()\{
		\FOR{all pixels $(i,j)$}
		\STATE $border_{ex,ij}\gets \frac{1}{3}\sqrt{\Sigma_c (border)_{ij,c}^2}\geq th_d$
		\STATE $body_{ex,ij}\gets \frac{1}{3}\sqrt{\Sigma_c (inner)_{ij,c}^2}\geq th_l$
		\STATE $body_{ex,ij}\gets body_{ex,ij} *(1-border_{ex,ij})$
		\STATE $heatmap_{ij}\gets border_{ex,ij}*0.9 + body_{ex,ij}*0.4$
		\ENDFOR	
\STATE \}
\end{algorithmic}
\end{algorithm}

\begin{algorithm}
\begin{algorithmic}
\caption{build\_ccell\_body($x_0,y_0,r,t,t_b$ etc) for type 1 or 2 cell, see fig. \ref{fig:maingallery}. *More descriptions in appendix}\label{buildccell}
\STATE build\_basic\_ball\_body($x_0,y_0,r,t$ etc)
\STATE $x,y\gets x+noise, y+ noise$
\STATE create\_skeleton()\{
  \bindent
  \STATE $bar_x=(x-x_0\leq r)*(x-x_0\geq -r)$
  \STATE $bar_y=(y-y_0\leq t_{b}/2)*(y-y_0\geq -t_b/2)$
  \STATE $bar=bar_x*bar_y$
  \IF{type 2}
    \STATE $pole_{y1}=(y-y_0\leq r*v_{s})$
    \STATE $pole_{y2}=(y-y_0\geq r*v_{s})$
    \STATE $pole_{x1}=(x-x_0\leq t_p/2)$
    \STATE $pole_{x2}=(x-x_0\geq -t_p/2$
    \STATE $pole= pole_{y1}*pole_{y2}*pole_{x1}*pole_{x2}$
    \STATE $ pole_{pos}=(pole\ge 0)*(1 - bar\ge0)$
    \STATE $ bar=bar+pole*pole_{pos}$
  \ENDIF
  \eindent
\STATE \}
\STATE rotate ball and bar by \(\theta\)
\STATE shift center of ball and bar to $(x_0,y_0)$
\STATE make\_explanation()

\end{algorithmic}
\end{algorithm}

We provide algorithms that can generate dataset on demand, illustrated in fig. \ref{fig:workflow}(B). Sample data can be seen in fig. \ref{fig:maingallery}, where the top three rows show the image data and the last three rows are the corresponding ground-truth heatmaps. Ten different classes of cells are shown along the columns. Types 0,1,2 are circular cells with border (algo. \ref{buildbasicball}), with a bar (or minus sign) and with a plus sign (algo. \ref{buildccell}) respectively. Types 3,4,5 are rectangular cells with different dominant colors. Types 6,7,8 are circular cells with one, three and eight tails respectively. The last class does not contain any cell. Three types of backgrounds are given to increase the variation of dataset, as shown separately in the first three rows of the same figure.

The algorithms are generally designed with simple formulas that correspond to mathematically well-defined shapes, as illustrated in fig. \ref{fig:workflow}(B). For example, in Algorithm 1, type 0 cell is a simple coloured elliptical cell with border coloured differently. We can see the cell body reflected as \(innerball=d<r-t\), describing the inner part of a circle. In algorithm 2, bar and pole reflect the shapes of minus and plus inside cells type 1 and 2. Noises etc will distort the shape further while retaining their main features. In both algorithm 1 and 2, make\textunderscore explanation() function is defined to highlight the features (dark red) and the localization (light red). They are easily obtained since the cells are obtained part by part.

The ground-truth heatmaps \(h_0\) have been designed to mark features that distinguish all the classes in a way that is as unambiguous as possible, subject to human judgment. Admittedly, there may not exist a unique unambiguous way of defining them. Where appropriate, the heatmaps could be readjusted by editing the heatmap generator classes in the package of codes. The heatmaps are shown in fig. \ref{fig:maingallery} row 4 to 6. With this dataset, fair comparison between heatmaps generated by different XAI methods can be performed. In this particular implementation, each ground-truth heatmap \(h_0\) is normalized to \([-1,1]\), thus heatmaps to be compared to \(h_0\) are expected to be normalized to \([-1,1]\) as well. Each \(h_0\) consists of an array of values of size (H, W) with three distinct regions (1) regions with value \(0\) (shown as white background) corresponding to regions that should not contribute to the neural network prediction, or non-features (2) regions with value \(0.4\) corresponding to localization (shown as light red region) and (3) regions with value \(0.9\) to delineate discriminative feature (shown as dark red region), where discriminative feature is also qualitatively considered part of localization. In this paper, two other distinct regions are defined symmetrically. They are \(-0.4\) and \(-0.9\) regions, to accommodate the fact that some heatmap methods have been interpreted in such a way that negative regions (shown as blue color in this paper) are considered as regions contributing \textit{against} a given decision or prediction \cite{7552539}. For our dataset, the ground-truth does not contain any such information that contributes negatively, although, as will be shown later, some XAI methods still do generate negative values.

For this paper, training, validation and evaluation datasets are prepared in 32, 8 and 8 shards respectively, each shard containing 200 samples uniform randomly drawn from the 10 classes. In another words, in total, the datasets contain 6400, 1600 and 1600 samples respectively. The dataset is prepared in shards for practical purposes, for example, to prevent full restart in case of interruption of data downloading and caching, and to facilitate more efficient process of training in evaluation mode indicated in fig. \ref{fig:workflow}(A) as part 2.

\subsection{Training}

\begin{figure*}
\centerline{\includegraphics[width=\textwidth]{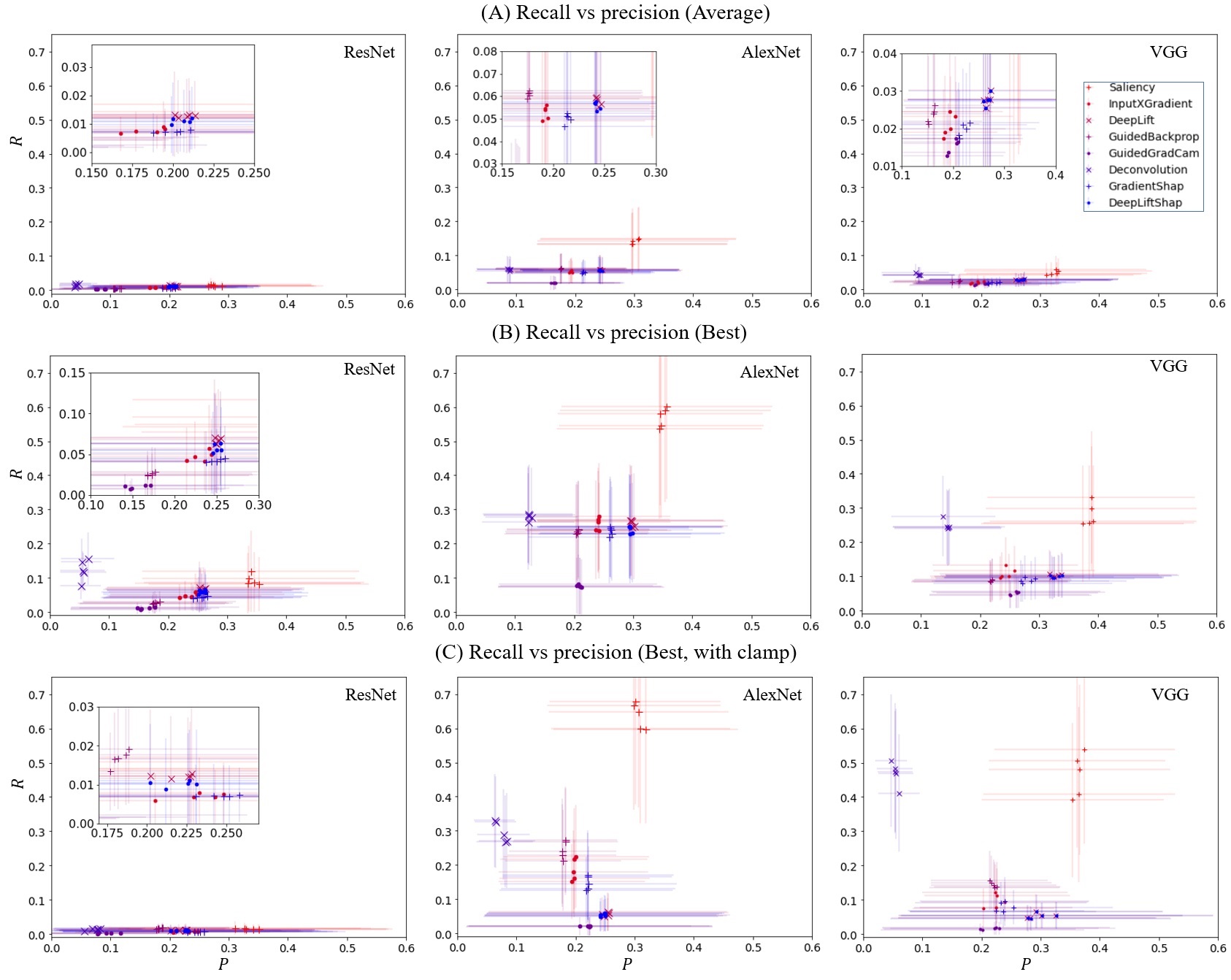}}
\caption{Recall and precision scores of five-band stratified heatmaps compared to ground-truth for ResNet, AlexNet and VGG and 8 different XAI methods. For all, higher recall and precision values are better, i.e. points located towards top-right are better. (A) Average and (B) maximum values of recalls and precisions (over soft five-band thresholds) of each sample of evaluation dataset are collected, then averaged over all these samples to be shown as individual points \((P,R)\) in the plot. (C) is the same as (B), except the scores are obtained after clamping process. Vertical and horizontal bars are the standard deviations over all samples of evaluation dataset correspondingly. Insets show zoom-in on selected regions. }
\label{fig:recall_vs_prec}
\end{figure*}

\begin{figure*}
\centerline{\includegraphics[width=0.9\textwidth, ]{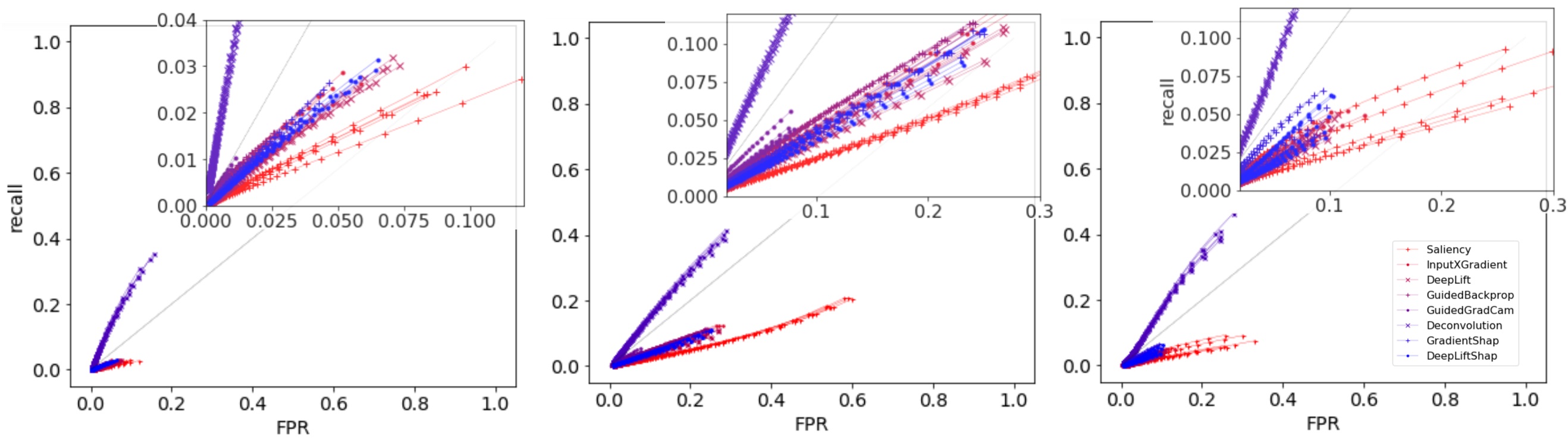}}
\caption{ROC curve for ResNet (left), AlexNet (middle) and VGG (right) for 8 different saliency methods (see legend, bottom-right), each XAI method applied on test datasets separately for 5 branch models. Both FPR and recall do not necessarily reach 1.0 as the thresholds are adjusted in multi-dimensional space. Most methods under our the experimental conditions specified here lie under traditionally poor ROC region.}
\label{fig:roc}
\end{figure*}

After the data is cached or saved, the process starts with \textit{training in continuous mode}, indicated as part 1 of the workflow fig. \ref{fig:workflow}(A). In this mode, pre-trained models are first downloaded from Torchvision and modified for compatibility with pytroch Captum API. The three pre-trained models used are AlexNet \cite{DBLP:journals/corr/Krizhevsky14}, ResNet34 \cite{7780459} and VGG \cite{7486599}, corresponding to workflow 1, 2 and 3 in the codes. In this phase, training proceeds continuously for the purpose of fine-tuning the models to our current data. The number of epochs and batch size are specified in table \ref{table:training}. Adam optimizer is used with learning rate \(lr=0.001\) for ResNet but \(lr=0.0001\) for AlexNet and VGG, and the same weight decay \(10^{-5}\) is used for all. Plot of losses against training iterations (not shown in this paper) is saved as a figure in part 1.2 of the workflow. We refer to the model trained after this phase as the \textit{base model}.

The next phase is the \textit{training in regular evaluation mode}, indicated as part 2 in fig. \ref{fig:workflow}(A). The training uses the same optimizer as the previous phase, and the number of epochs used are also shown in table \ref{table:training}. Evaluation is performed every 4 training iterations; more accurately, this part is known as validation in machine learning community, separate from the final evaluation. Each validation is performed on a shard randomly drawn from the 8 shards of the validation dataset. We set the \textit{target accuracy} to 0.96. If during validation, the accuracy computed on that single shard exceeds the target accuracy, the training is stopped and evaluation on all validation data shards is performed. The total validation accuracy is used to ensure that the validation accuracy on a single shard is not high by pure chance. While the total validation accuracy can be slightly lower, our experiments so far indicate that there is no such problem. Furthermore, only ResNet attained the target accuracy within the specified setting. For AlexNet and VGG, 0.96 is not exceeded throughout and early stopping mechanism is triggered to prevent unnecessarily long, unfruitful training; note that, fortunately, the total accuracy when evaluated on the final evaluation dataset is still very high, as shown in table \ref{table:training}. The early stopping mechanism is as the following. Whenever validation on a single shard does not achieve the target accuracy but (1) if there is no improvement in the validation accuracy, then early stopping counter \(n_{early}\) is increased by one (2) if there is improvement in the validation accuracy, then \(n_{early}\to n_{early}\times r_f\), where \(r_f< 1\) is the refresh fraction, so that the process is given more chance to train longer. If \(n_{early}\) becomes equal to the early stopping limit, \(n_l\), training is stopped. 

\begin{table}[htbp]
\centering
\caption{Training settings and performances on the ten-classes data. TC and TR denote trainings in continuous and regular evaluation mode respectively. \(<Acc.>\) denotes average accuracy over 5 models branched from the base model. \(n_{batch}\) is the batch size, \(n_{ep}\) no. of epochs. Image shapes are \((H,W)\) where \(H=W\).}
\scalebox{0.9}{
\begin{tabular}{c|c|c|c|c|c|c|c}
\hline
& & & TC & \multicolumn{3}{c|}{TR} & Eval \\
\hline
& H,W & \(n_{batch}\) & \(n_{ep}\) & \(n_{ep}\) & \(n_l\) & \(r_f\) & \(<Acc.>\) \\
\hline
ResNet & 512 & 4 & 2 & 4 & 24 & 0.4 & 0.951\\

AlexNet & 224 & 16 & 16 & 16 & 48 & 0.3 & 0.980 \\

VGG & 224 & 16 & 4 & 4 & 48 & 0.3 & 0.986 \\
\hline
\end{tabular}
}
\label{table:training}
\end{table}

We repeat the above process of training in regular evaluation mode 4 other times starting from the base model, and thus we have a total of 5 branch models. Note that \(r_f\) and \(n_l\) are set so that AlexNet and VGG can be trained for longer period (shown in table \ref{table:training}), since they both achieve lower accuracy performance than ResNet if given the same number of epochs, \(n_l\) and \(r_f\). This is possibly because (1) larger batch size means fewer iterations per epoch and (2) improvement in accuracy is inherently slower, considering that ResNet has been known to generally perform better. Here, comparing accuracy of prediction in a precise manner is not very meaningful, as we are focusing on the heatmaps later. Perfect accuracy is not attained and there is no guarantee that DNN will ever attain a perfect performance. However, we take this opportunity to compare heatmaps between correct and wrong predictions (see discussion section later). Finally, note that there is no need for k-fold validation here since the validation dataset is completely separate from the training dataset.

\subsection{Evaluation and XAI implementation}
\label{subsection:eval}
This part corresponds to part 3 of fig. \ref{fig:workflow}(A), where heatmaps \(h\) are computed using the following XAI methods available in pytorch Captum API: Saliency \cite{Simonyan14a}, Input*Gradient \cite{Shrikumar2016NotJA} (reference based on Pytorch Captum's page), DeepLift \cite{pmlr-v70-shrikumar17a}, GuidedBackprop \cite{springenberg2014striving}, GuidedGradCam \cite{GradCAM}, Deconvolution \cite{10.1007/978-3-319-10590-1_53}, GradientShap \cite{NIPS2017_7062}, DeepLiftShap \cite{NIPS2017_7062}. Integrated-Gradients \cite{10.5555/3305890.3306024} has been excluded as it is comparatively inefficient with ResNet. Note also that the original implementation of \(\epsilon\)-Layerwise Relevance Propagation (LRP)\footnote{We thank Leon Sixt for the relevant comment.} \cite{PixelWiseLRP} has been shown to be equivalent to gradient*input or DeepLIFT depending on a few conditions \cite{pmlr-v70-shrikumar17a}. For all heatmaps, we compute the heatmaps derived from the predicted values, not the true values (for some XAI methods, explanation can be extracted from the probability of predicting not only the correct class, but also other classes). The following is the sequence of processing leading to the final results.

\textit{Channel adjustments}. Each heatmap \(h\), which has (C, H, W) shape (C=3 for 3 color channels), is compressed along the channels to (H, W) by \textit{sum-pixel-over-channels}, where the values are summed pixel-wise along all channel, i.e. \(s_c (h_{ij})=\Sigma_{c=1,2,3} h_{ij, c}\) when written component-wise. This is so that it can be compared with \(h_0\) of shape (H, W). Normalization to \([-1,1]\) is also performed in the following \textit{channel adjustment} process: \(h \to s_c(h) \to h/max(|h|)\). \(max(|h|)\) is the maximum absolute value over all pixels in that single heatmap. The practice of summing over channels can be seen, for example, in LRP tutorial site \cite{LRPTutorial} and \cite{7552539}.

\textit{Five-band stratification}. Adjusted heatmaps \(h\) will subsequently be evaluated using \textit{five-band score}, where each pixel needs to be assigned one of the five values $0.9,0.4,0,-0.4,-0.9$ described in section \ref{subsection:data}. The value 2 is designated for discriminative feature, 1 for localization, 0 for irrelevant background, while -1 and -2 are symmetrically defined for negative contribution to model prediction or decision. Recall that our ground-truth heatmaps \(h_0\) pixels have been assigned one of the following values 0, 0.4 and 0.9. Regardless of the intermediate processing of the heatmap \(h\), the mapping for \(h_0\) is always such that \(0.9\to 2\), \(0.4\to 1\) and \(0\). To map \(h\), which has been normalized to \([-1,1]\) by now, a threshold of the form \(t=[-t_{(2)},-t_{(1)},t_{(1)},t_{(2)}]\) is used, so that for each pixel \(h_{ij}\), a transformation we refer to as \textit{five-band stratification} is performed in the following manner: \(h_{ij}\to 2\) if \(h_{ij}> t_{(2)}\), to \(1\) if \(h_{ij}\in (t_{(1)}, t_{(2)}]\), to \(0\) if \(h_{ij}\in (-t_{(1)}, t_{(1)}]\), to \(-1\) if \(h_{ij}\in (-t_{(2)}, -t_{(1)}]\) and to \(-2\) if \(h_{ij} \le -t_{(2)}\). Bracketed sub-script here is used to denote the component of \(t\) if it is regarded as a vector for notational convenience later. Up to this point, we have \(S^5(h_0), S^5_t (h)\) where \(S^5\) denotes the five-band stratification. 

\textit{Five-band score}. After stratification, for each heatmap, we compute accuracy \(A\), precision \(P=\frac{TP}{TP+FP+\epsilon}\), \(R=\frac{TP}{TP+FN+\epsilon}\), where accuracy is the fraction of correctly assigned \(h_{ij}\) pixel over the total number of pixels, TP is the number of true positive pixels, FP false positives, FN false negatives and \(\epsilon=10^{-6}\) for smoothing$  $. TP is slightly different from TP used in binary case. We only count \(TP\) when \(h_{0, ij}\neq 0\) and \(h_{ij}=h_{0,ij}\), i.e. we use the stringent condition where the labels for localization and features must be correctly hit to achieve a true positive. Likewise, \(FP\) is counted when \(h_{0,ij}=0\) and \(h_{ij}\neq 0\) plus \(h_{0, ij}\neq 0\) and \(h_{ij}\neq h_{0,ij}\) whereas \(FN\) when \(h_{0,ij}\neq 0\) and \(h_{ij} = 0\). To plot receiver operating characteristics (ROC), false positive rate \(FPR=\frac{FP}{FP+TN+\epsilon}\) is also computed, where TN is the number of true negatives \(h_{ij}=h_{0,ij}=0\).

\begin{figure}
\centerline{\includegraphics[width=\columnwidth, ]{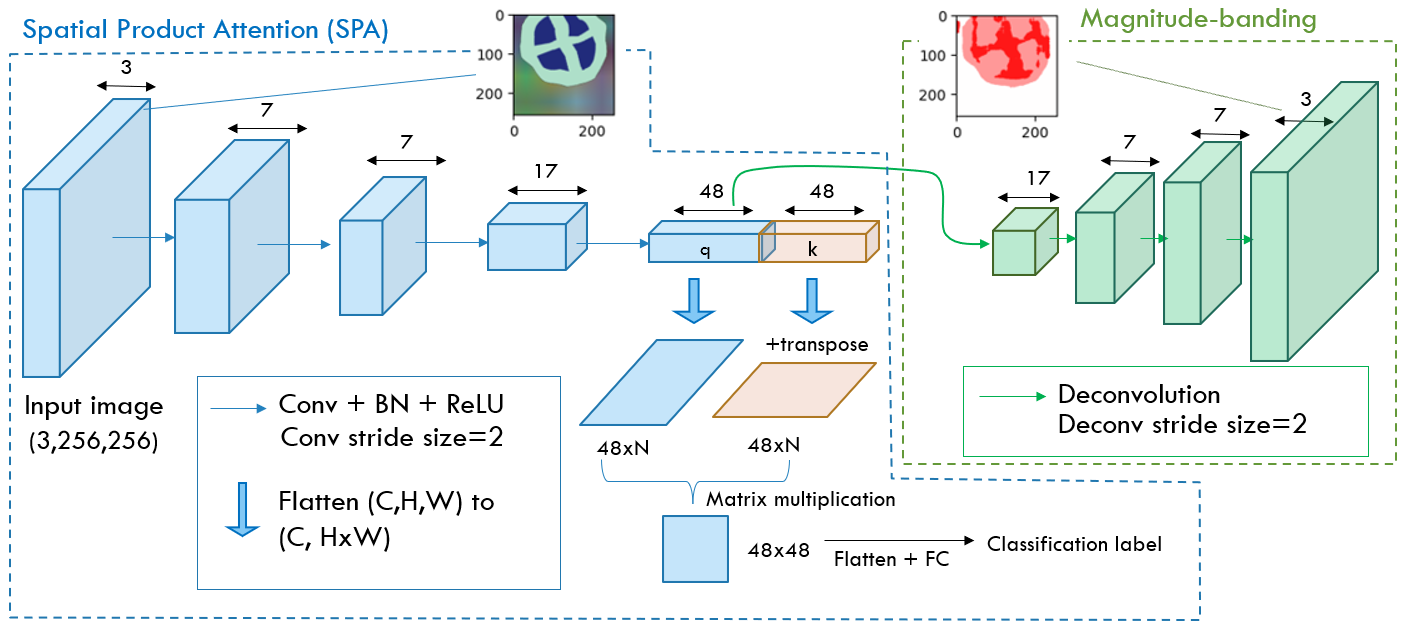}}
\caption{Spatial Product Attention (SPA) and Magnitude-banding. SPA is a classification model with query $q$ and key $k$ abstraction (like transformer). Magnitude-banding is achieved through deconvolution layers on query $q$ (green dotted box).}
\label{fig:mabspa}
\end{figure}

\textit{Soft five-band scores}. As seen, the threshold defined above is sharp, and the value near any of the thresholds \(\pm t_{(i)}\) might not be properly accounted for. We thus instead use soft five-band scores, where the metrics are collected for different thresholds. More precisely, for the k-th data sample, we obtain \((A,R,P)^{(k)}_{t_m}\) for \(t_m=[-0.5+md,-0.3+md,0.3-md,0.5-md]\) where \(d=0.005\), \(m=0,1,\cdots,n_{soft}\), and \(n_{soft}=55\) after comparing the stratified ground-truth \(S^5 (h_0)\) with \(S^5_{t_m}(h)\), where \(h\) has undergone \textit{channel adjustment} process previously described. The best and average values of \(X=A,R,P\) for sample \(k\) over the different thresholds, \(X^{(k)}_{avg} = \frac{1}{n_{soft}}\Sigma_{t_m} X^{(k)}_{t_m}\) and \(X^{(k)}_{best}= max_{t_m}\{X^{(k)}_{t_m}\}\) respectively, are then saved sample by sample into a csv file in the XAI result folder for analysis in the discussion section. These values are identified by their positions among the shards, the predicted class and the true class. Note: the choice of \(t_m\) is chosen such that localization is found at the ``middle range" between $\sim[0.3,0.5]$ and discrimination at the ``higher range" i.e. stronger signals.

\textit{Receiver operating characteristic}. To compare the performances of different XAI methods mentioned above, ROC is also obtained as shown in fig. \ref{fig:roc}. For each threshold \(t_m\), mean values of \(\{FPR^{(k)}_{t_m}\}\) and \(\{R^{(k)}_{t_m}\}\) over all samples in the evaluation datasets contribute to a single point in the figure. Unlike the usual binary ROC, changing thresholds in the multi-dimensional space we defined does not guarantee the change from \(FP\) to \(TP\) (or vice versa). For example, a point that begins as \(FN\) that predicts label \(0\) can become \(TP\) or \(FP\) if the true value are \(1\) and \(2\) respectively when the \(0\) thresholds are lowered. Hence, we will not always obtain a curve that starts with \((0,0)\) and ends with \((1,1)\) in the ROC space, unlike the usual ROC curve. Regardless, by simple understanding of rate of change of FPR and recall, the usual rule of thumb that assigns steeper increase in recall to better ROC quality should still hold. Mathematically, the more optimal ROC curve lies nearer the top-left vertices of the convex hull formed by the points. There has been studies on multi-dimensional ROC curve with its ``area under volume" \cite{Srinivasan99noteon, 10.1007/978-3-540-39857-8_12}, though the difficulty of observing them makes them unsuitable for visual comparison here. With the definition of TP, FP, TN, FN above, we have instead created pseudo-binary conditions.

\subsection{Magnitude-banded Class Activation Mapping}
\label{subsection:mabcam}
The central idea in this paper has been highlighted in the previous sub-section: the quality of XAI methods is measured in a straightforward manner, specifically by measuring pixelwise hit/miss against ground-truth heatmaps/attributions using metrics such as recall and precision. We have proposed five-band score to quantify different levels of explanation, and we have assigned $0.4$ as the value corresponding to localization and $0.9$ to more prominent, distinguishing features, i.e. pixelwise attribution is banded by magnitude. In this sub-section, we present the mabCAM (Magnitude-banded Class Activation Mapping), a general arrangement of deep neural network modules intended to produce high-quality magnitude-banded heatmaps.

Before going into details, we briefly describe the components used to generate mabCAM heatmaps. A series of deconvolution layers (denoted $mab$) will be applied to feature maps at the end of a convolution layer. In this paper, $mab$ is applied to feature maps from the last convolution layer, similar to the original CAM which computes the weighted sum of features in the last convolution layer. Our deconvolution layers are not attached to activation functions or batch normalization layers, and each layer has stride 2. The goal is to expand the dimensions of feature maps to approximately the same height and weight as the original input, just like the decoding part of a generic variational auto-encoder.  The number of channels in deconvolution layers correspond to the number of channels in the convolution layers preceding them (ideally, we wish to use as few as possible parameters just like CAM, but it did not work well enough). The final deconvolution layer has 3 channel outputs corresponding to 3 classes of labels: non-feature, localization and features, corresponding to heatmap values $0,0.4,0.9$ respectively. This arrangement allows optimization via the minimization of cross-entropy loss similar to semantic segmentation. 

Some remarks: (1) ideally, there are 5 channel outputs corresponding to $0.9,0.4,0,-0.4,-0.9$ rather than 3 channels, the two missing classes being the ``negative attributions". However, in our dataset, the design does not incorporate any features that contribute in a ``negative" way, not to mention the fact that it is not yet clear what role the negative attribution values play in existing XAI methods, at least w.r.t the negative values (blue patches) that we observe in heatmaps produced by these methods on our dataset. (2) Semantic segmentation can be very costly, especially when there are many classes. There need to be as many output channels as the class labels, and each ground-truth label may need to be marked manually. Our procedure automatically generates similar pixelwise label, which is convenient. Note that using our magnitude-banding framework on other datasets still requires some manual labeling of features and localization areas. Unlike semantic segmentation, our magnitude banding only requires 3 classes (at most 5, if the negative attribution is included), thus the manual labeling effort is not as taxing.

\textit{The mabCAM design}. The implementation in this paper is geared towards a small model with only $82376$ parameters (compared to ResNet50 with over 20 million parameters). Regardless, the idea is general and widely applicable, as the following. First, we introduce the Spatial Product Attention (SPA), as shown in fig. \ref{fig:mabspa}. SPA is a series of convolution layers set up to quickly downsize the feature maps, which are then split to yield query $q$ and key $k$ abstractions, like the attention mechanism in a transformer encoder \cite{NIPS2017_3f5ee243}, analogous to a retrieval system. In SPA, $q$ is designed as the components that will later be fine-tuned to produce our attribution heatmaps. For image classification, the key and query are flattened, followed by a matrix multiplication \(qk^T\), resulting in a 48 by 48 feature map (note that each of its pixel is already influenced by all pixels in the input image upstream). With this arrangement, the size of SPA's pre-FC feature map is independent of the input image size, as long as the input image size is not too small; thus we do not need global average pooling commonly used in a CNN. 

The component that generates mabCAM heatmap is $mab$, or the the magnitude-banding module, as seen in dotted green box of fig. \ref{fig:mabspa}. It is a series of deconvolution layers, and no special novelty is required here. Each layer also has stride=2, resulting in heatmap dimensions approximately equal to the input image. As a whole, \(y,h=mabCAM(x)\) where \(x\) is the input image, \(y\) the raw predicted class label and \(h\) the raw predicted heatmap of shape \((3,h',w')\). The predicted feature maps is computed by, first, \(h_1=argmax(h, dim=0)\) and finally \(attr=(h1==1)*0.4+(h1==2)*0.9\). To train the model, cross entropy loss is minimized for both class labels and magnitude-banding labels, i.e. \(loss=CELoss(y,y0)+\lambda_{mab}CELoss(h,h0)\). Note: magnitude-banding labels mapping is $0\rightarrow 0, 0.4\rightarrow 1, 0.9\rightarrow 2$. We use Adam optimizer with learning rate \(0.001\) and \(\beta=(0.5,0.999)\), batch size \(16\) and \(mab\) regularization \(\lambda_{mab}=5\). No special hyperparameter tuning is performed. Training data consists of 4096 samples uniformly drawn from the 10 classes of the synthetic cells previously described, while validation and test data consist of 1024 samples each. Additionally, smaller 3-class demo is also available in the appendix. We allow training to last for at most 128 epochs, though the training is terminated if the validation accuracy reaches 0.975. Once training is completed, five-band scores  are computed similar to the way we compute them for other XAI methods. The difference lies in the soft thresholding: since mabCAM is designed to be compatible with the discrete heatmap values, we do not need to perform soft-thresholding across 56 soft-steps. The results from mabCAM will be discussed at the end of the next section.

\section{Discussion}
\label{section:discussion}
\subsection{Recall vs Precision}
We provide recall vs precision scores as shown in fig. \ref{fig:recall_vs_prec}. Each point in the plot corresponds to an XAI method, for example Saliency, applied on a single branch of the corresponding model trained from the base model. There are 5 points per method as we have trained 5 branches per architecture. Naturally, the higher \(P\) and \(R\) are, the better is the XAI method. Each point can be denoted by \((R_{stat},P_{stat})\), where \(X_{stat}=\frac{1}{N}\Sigma_{k=1}^N X^{(k)}_{stat}\), \(X=R,P\), \(stat=avg,best\) and \(N\) is the number of data sample in the evaluation dataset. Thus fig. \ref{fig:recall_vs_prec}(A) is a plot of \(R_{avg}\) vs \(P_{avg}\) for ResNet, AlexNet and VGG respectively. Likewise, fig. \ref{fig:recall_vs_prec}(B) is \(R_{best}\) vs \(P_{best}\). After qualitative assessment of some of the generated heatmaps, we perform similar analysis by applying clamping to heatmap values after the first normalization process, following roughly the idea in \cite{tjoa2019enhancing}. In other words, the \textit{channel adjustment} process described in the previous section is changed to \(h\to h/max(|h|) \to C_{[c_1,c_2]}(h) \to s_c(h) \to h/max(|h|)\) where \(C_{[c_1,c_2]}(h_{ij})=c_2\) if \(h_{ij}\ge c_2\), \(C_{[c_1,c_2]}(h_{ij})=c_1\) if \(h_{ij}\le c_1\) and otherwise \(C_{[c_1,c_2]}(h_{ij})=h_{ij}\). A different set of soft thresholds has been used to match the clamping process as well, with \(t_m=[-0.9+md,-0.5+md,0.5-md,0.9-md]\) where \(d=0.01\), \(m=0,1,\cdots,n_{soft}\), and \(n_{soft}=40\) with the clamping threshold given by \([c_1,c_2]=[-0.1,0.1]\). Fig. \ref{fig:recall_vs_prec}(C) is thus the same as fig. \ref{fig:recall_vs_prec}(B), except with clamping process applied, where we do observe some changes in the precision and recall scores, more notably for AlexNet and VGG, though not necessarily better.

\begin{figure}[htpb]
\centerline{\includegraphics[width=\columnwidth, ]{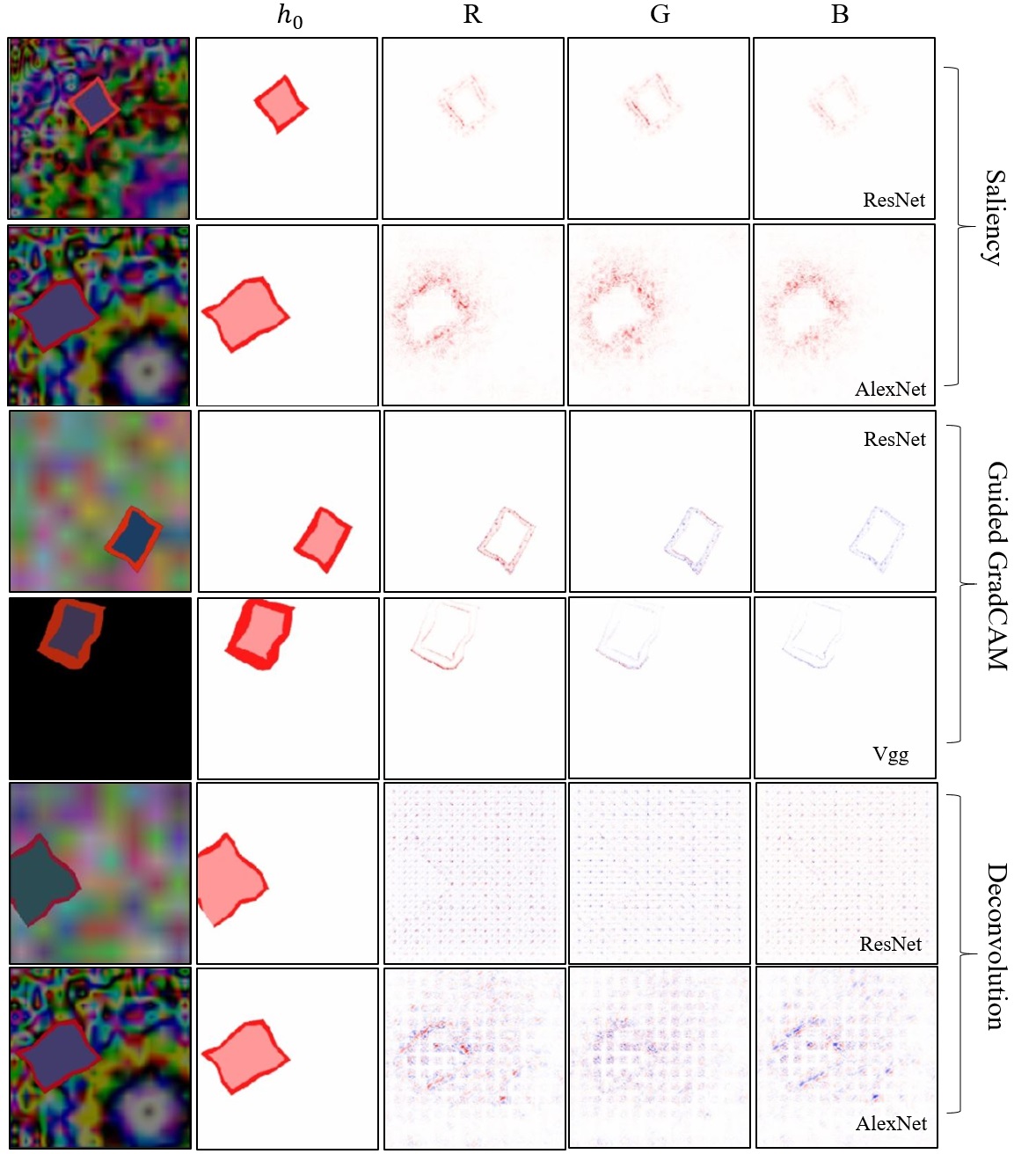}}
\caption{Visual comparison of heatmaps generated by Saliency, Guided GradCAM and Deconvolution. Different color-channel responses are shown under R, G and B columns respectively, with the original image in the left-most column and ground-truth \(h_0\) in the second left-most column. All heatmaps above are obtained from correctly predicted samples.}
\label{fig:heatmap1}
\end{figure}

\begin{figure}[htpb]
\centerline{\includegraphics[width=\columnwidth, ]{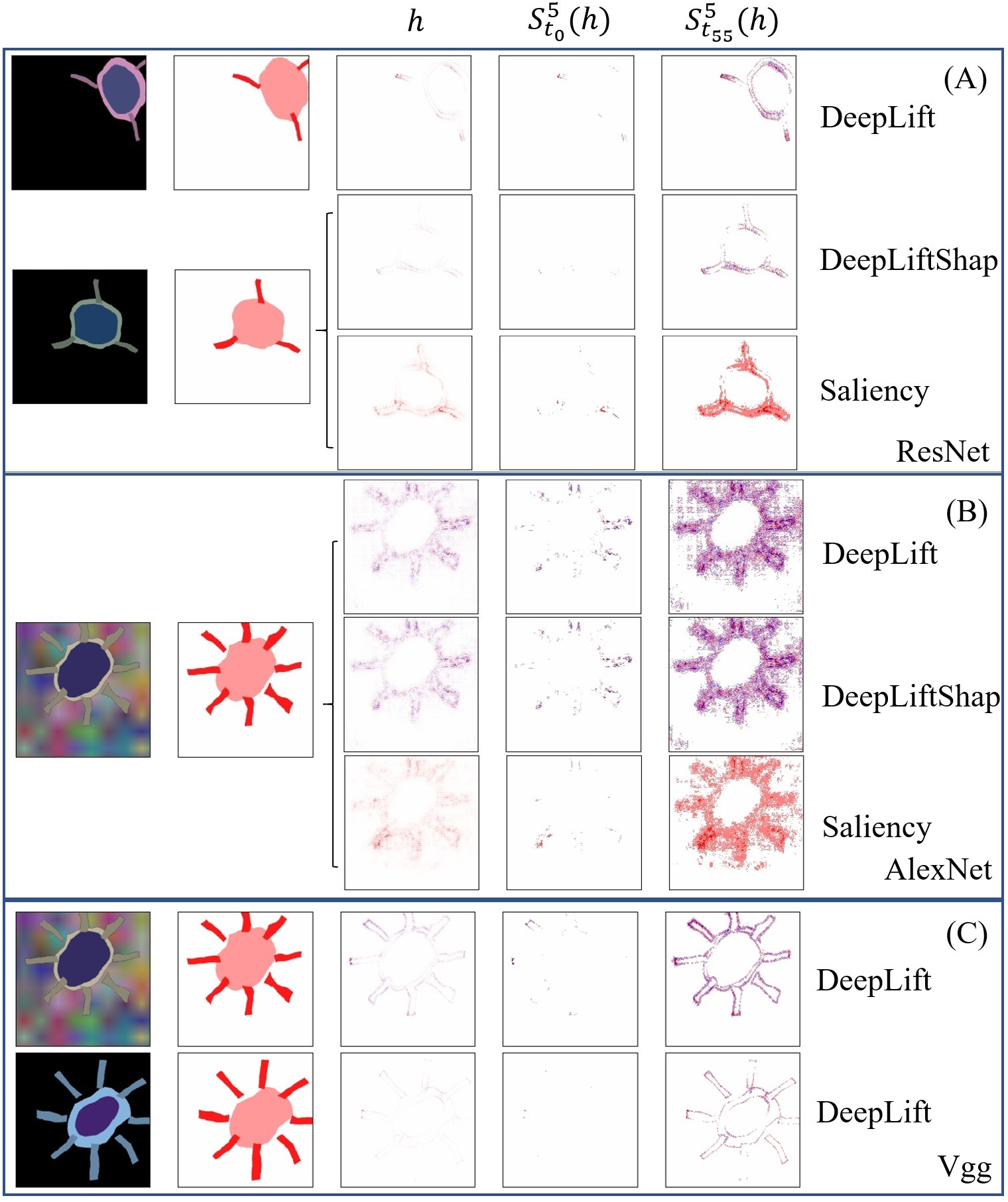}}
\caption{Visual comparison of heatmaps generated by DeepLift and DeepLiftShap, with Saliency for comparison. Column \(h\) is obtained after \textit{summing pixel over channels}. Columns \(S_{t_0}^5(h)\) and \(S_{t_{55}}^5(h)\) are obtained after five-band stratification using the first and last thresholding described in section \ref{subsection:eval}, where \(t_0=[-0.5,-0.3,0.3,0.5]\) and \(t_{55}=[-0.225, -0.025, 0.025, 0.225]\). (A) Visualization of how DeepLift and DeepLiftShap on ResNet generally score slightly lower in recall scores than Saliency (B) DeepLiftShap and DeepLift appear to produce similar heatmaps for VGG and ResNet, though the SHAP variant appears to remove some artifacts in Alexnet. Negative attribution values (blue) are difficult to interpret. (C) Heatmaps for correct prediction of cells from the same class, cell type 8, generated using DeepLift applied on VGG. See more in appendix.}
\label{fig:heatmap2}
\end{figure} 

Recall scores are generally low in most of the points in fig. \ref{fig:recall_vs_prec}, indicating high FN. The first obvious cause is the fact that most XAI methods in all architectures appear to assign 0 values to regions that contain either localization pixels or discriminative features. For example, fig. \ref{fig:heatmap1} shows the heatmaps from different channels R, G and B (extracted before summing pixel over channels). The heatmaps generally appear granular and non-continuous, having many white pixels in between the red pixels, thus contributing to false negatives. Furthermore, most of the inner body of the cells (represented by light red color in the ground-truth \(h_0\)) is completely unmarked by most of the XAI methods, contributing to very large amount of false negatives. The highest recall values in fig. \ref{fig:recall_vs_prec}(A) are attained by Saliency applied on AlexNet. This is consistent with visual inspection of the heatmaps across different methods and architectures, because Saliency assigns a lot more red pixels in relevant regions while other methods often assign blue pixels (negative values) in unpredictable manner and highlight only the edges.

Similar to the heatmaps shown in fig. \ref{fig:heatmap1} produced by Guided GradCAM applied on VGG, many of the XAI methods only highlight the edges of the cell borders, sometimes faintly. As such, comparatively high recall values for Saliency can be qualitatively accounted for by the halo of high-valued heatmap encompassing the relevant area, although not in a very precise and compact manner. Deconvolution, on the other hand, has relatively higher recall scores due to the large amount of artifact pixels. The quality of its heatmaps has been therefore undermined, reflected as low precision score. Other methods such as guided GradCAM are still capable of highlighting some of the relevant regions, and, to reiterate, many of them tend to highlight the edges as seen in the heatmaps in the supp. material. Also, AlexNet tends to produce denser heatmaps than the other two, giving rise to slightly higher recall scores than VGG, while the average scores for ResNet are very low. Depending on the context, different XAI methods can be the better choice based on their strength and weaknesses. More ideally, existing interpretations may need to be defined w.r.t context of the problems solved by the particular DNN e.g. when DNN performs image classification, users need to be informed whether localization is considered ``part of explanation" or specific parts of an image contributes to the explanation. 

Differences in responses to color channels are also observed. Saliency method appears as positive values (red) in all channels as shown in fig. \ref{fig:heatmap1}, although type 3 cell has only 1 color channel whose input signal is strong because it has border whose pre-dominant color is red; in the implementation, the normalized border color is roughly \((0.8,0.1,0.1)\) with small uniform random perturbation. On the other hand, Guided GradCAM marks green and blue channels with negative values. If they are to be interpreted as negative contribution, the interpretation will be consistent. But when the heatmaps are summed over channels as we have done, the offsetting effect of the negative values become questionable. In other methods, such color responses are variable. For example, input*gradient for AlexNet do appear to exhibit color responses as well (not shown), although the quality is highly variable too. It is thus difficult to strongly recommend any one method specializing on color-detection, even for guided GradCAM.

\textit{Interpretation of heatmap values}. Fig. \ref{fig:heatmap2} shows in column \(h\) the heatmaps obtained after \textit{summing pixel over channels}, one of the earlier processes in the previous section.  The figure shows the effect of soft five-band stratification as well, which demonstrates that the appropriate selection thresholding does affect the scores. In previous section, we addressed this by distinguishing between the best and average of recall and precision values over the soft thresholds, which is the main purpose of fig. \ref{fig:recall_vs_prec}(B). The effect of threshold change is variable across different XAI methods. If we focus on recall scores, from visual inspection of fig. \ref{fig:heatmap2}(B), the XAI community may need to revise the idea of negative values in heatmaps. Clearly, DeepLift and DeepLiftShap examples show that they will score much better recall if we take the absolute values of the heatmaps and apply the same process from stratification to the computation of five-band scores. 

\textit{SHAP and DeepLift}. When SHAP is applied to DeepLift, the effect appears to be background artifact removals, thus confining non-zero heatmap pixel values to more relevant regions (fig. \ref{fig:heatmap2}(B) and supp. materials). Blue pixels (negative values) mark some of the correct areas that we regard as discriminative features, but interpreting the blue pixels for these methods as negative contribution seems to be inappropriate. Applying absolute value to the negative pixels may improve its recall scores etc; generally, more investigation on the signals activated by the background may be necessary. Furthermore, we need to point out that heatmap shapes could be inconsistent even if the same classes of cells are correctly predicted, as shown in fig. \ref{fig:heatmap2}(C). The figure shows two heatmaps of different qualities generated by DeepLift for VGG for cell type 8 that are correctly predicted. It may be tempting to make guesses regarding possible reasons, for example, backgrounds might play a large role in distorting the heatmaps. 

From observing fig. \ref{fig:recall_vs_prec} and many heatmaps, for examples the figures in the appendix, it is tempting to deduce that deeper networks (AlexnNet shallowest, followed by VGG, then ResNet deepest) tend to produce heatmaps that are more sensitive to the edges but cover less thoroughly the bulk of discriminative features and localization regions. To test this, we conduct a test on AlexNet modified by systematically adding more and more convolutional layers, trained and then evaluated for its precision vs recall in the same manner as before. The number of layers added are 1, 2, ..., 8, and the plot is made by computing mean values of recall and precision like before, except that the points are collected separately based on the predicted values (whereas in the previous section, averages are taken over all test samples regardless of predicted values). The expectation is for the precision and recall values to be nearer to 1 (more towards top right of the plot) for the modified AlexNet with less additional layers. However, as shown in appendix fig. \ref{fig:appdx1}, this does not appear to be the case.

\textbf{ROC curve}. ROC plots in fig. \ref{fig:roc} show that most heatmap methods tested lie on traditionally poor ROC regions. There appears to be trade-offs between higher recall values (which is good) and higher FPR (which is bad), most prominently shown by Saliency method. Deconvolution appears to be the best, as it has the greatest rate of increasing recall compared to FPR. However, this is misleading, since deconvolution starts with many FP predictions in all three architecture, as shown by the grid-like artifacts in fig. \ref{fig:heatmap1}. This causes FP to change more quickly, and the ROC fails to make good comparison between deconvolution and other methods. Saliency tends to ``over-assign" the heatmap pixels around the correct region; consider fig. \ref{fig:heatmap1}, \ref{fig:heatmap2} and appendix, compare it to, for example, Guided GradCAM, DeepLift. Unlike DeepLift and DeepLiftShap, Saliency ROC shows higher recall because of more correct assignments of positive (red) values, but also higher FPR because of the assignment of positive values in supposedly white regions. Guided GradCAM appears to have some difficulty improving through the change of thresholds. Considering the way \textit{sum-pixel-over-channels} is performed and its color-channel sensitivity, its performance might have suffered through incompatible heatmap pre-processing. ROC for other methods do not provide sufficiently distinct trends that favor the adoption of one method over another. There may be a need to investigate the different ways soft-thresholding can be performed for specific XAI methods to at least bring the ROC curves to traditionally favorable regions.

\begin{figure}[htpb]
\centerline{\includegraphics[width=\columnwidth]{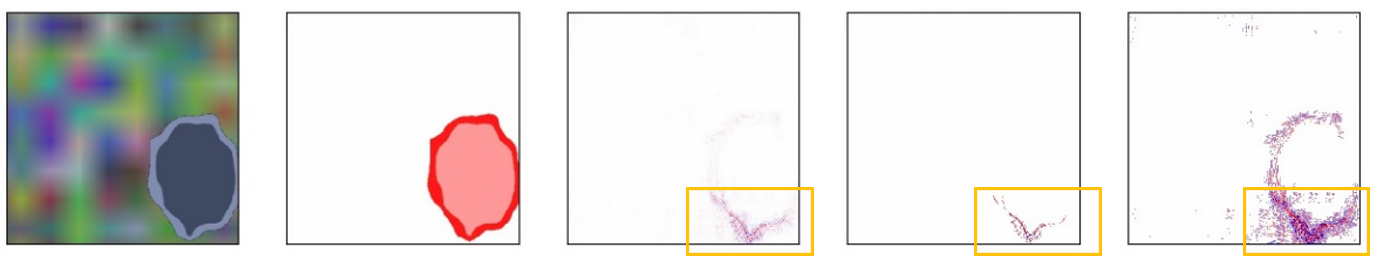}}
\caption{One example from DeepLift with wrong prediction (from appendix fig. 16).}
\label{fig:heatmapwrong}
\end{figure}

\textbf{Other observations}. For images of type 9 (no cell present), generally we see heatmaps in the form of artifacts appearing as well-spaced spots, forming lattice (see appendix fig. 40 etc), similar to the heatmaps from deconvolution method in fig. \ref{fig:heatmap1}. In some cases, for example see orange boxes in fig. \ref{fig:heatmapwrong} (from appendix fig. 16 row 2), we can see that DeepLIFT is able to ``provide" the correct reasoning for wrong prediction. In that figure, type 2 cell that has shape that almost looked like a single tail was mistaken as type 6, though some similar wrong predictions are not highlighted in similar manner. In many other cases of wrong predictions (see the heatmaps in the appendix), it is unclear what the highlighted regions mean.

\subsection{Results with mabCAM}
Here we discuss the main results obtained from the 10-class experiments stored in checkpoint/project02 folder (to find it, follow the instructions in our updated github) while all other results from experiments with the smaller 3-class data will be briefly discussed in the appendix. Training lasted the entire 128 epochs without early termination, reaching validation accuracy of 92.19\% at epoch 121 (starting at epoch 0). The test accuracy is 90.04\%. 

\begin{figure}[htpb]
\centerline{\includegraphics[width=\columnwidth]{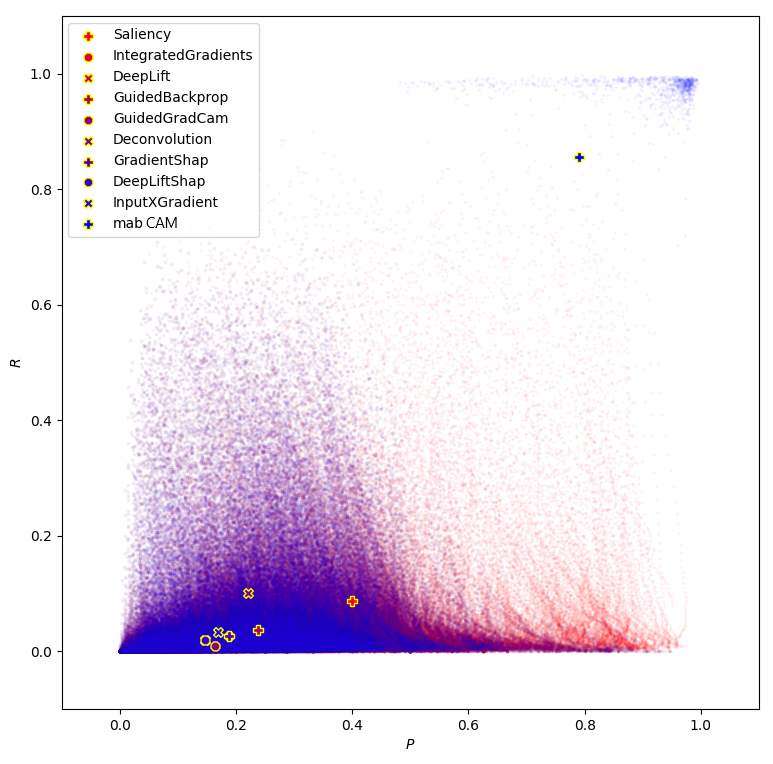}}
\caption{Recall versus precision scores of five-band stratified heatmaps compared to ground-truth for mabCAM. Unlike previous figures, we scatter the scores of all individual test samples in the background. The mean values are presented as opaque coloured marker with yellow border highlight. \textit{Zoomed-in view on an electronic media is highly recommended.}}
\label{fig:mabcam_main_result}
\end{figure}

\begin{figure}[htpb]
\centerline{\includegraphics[width=1\columnwidth]{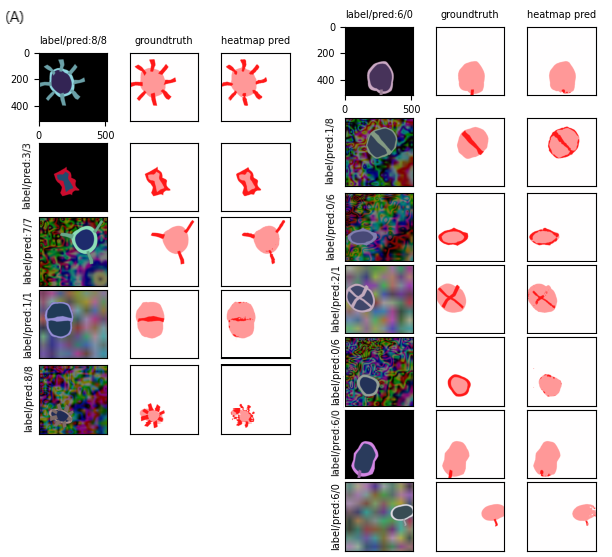}}
\caption{(A) From left to right: cell images, ground-truth and predicted magnitude-banded heatmaps. All labels are correctly predicted. (B) similar to (A), except all labels are wrongly predicted.}
\label{fig:mabcam_gallery_correct}
\end{figure}

As expected, since mabCAM is finetuned against ground-truth heatmaps, it produces significantly better recall and precision compared to all other XAI methods, as shown in fig. \ref{fig:mabcam_main_result}(A). Clearly, our magnitude-banded heatmaps ($mab$) show much higher recall and precision scores with mean values beyond $0.8$ for both, while other methods at most barely crossed $0.4$ precision. Some samples of heatmaps generated by mabCAM are shown in fig. \ref{fig:mabcam_gallery_correct}, demonstrating high quality (although imperfect) heatmaps. This is useful because it demonstrates that explanations can be tailored to the context, especially by matching them to ground-truth user-defined explanations. Naturally, users will have to define what is considered the suitable explanation beforehand, and a few-shot training on explanation data is a possible follow up for future works. This framework ensures that no ambiguity will arise, assuming that users agree on the pre-defined explanations.

Finally, we collected some samples of wrongly predicted images as shown in fig. \ref{fig:mabcam_main_result}(B). We see that mabCAM confuses class 0 and 6 often, which is not surprising, given that they differ by a single tail that sometimes blend into the noisy background colour or is too small. Other samples also demonstrate how background noise colours might be mistaken as tails e.g. second row of fig. \ref{fig:mabcam_gallery_correct}(B) highlights the edge of the cell even though there is no tail. The heatmaps with wrong classifications generally show inconsistency, where features (patches with value $0.9$, dark red) are mistakenly served as explanations.

\section{Conclusion}
\noindent \textit{Recommendations and Caveats}. Regardless of the imperfect performance, relative comparisons between the XAI methods can be made. 
\begin{itemize}
\item Saliency method appears to highlight the relevant regions in the most conservative way, which is more suitable for localization in the case false positives are not important. In particular, AlexNet is scoring the highest recall.
\item If only the edges of features are needed, then VGG and ResNet with input*grad, DeepLift, DeepLiftShap seem to be the reasonable choices. The same heatmap methods for Alexnet seem to produce heatmaps that go beyond capturing just the edges in rather inconsistent ways. Compared to Saliency, these methods may be more useful to detect small, hard to observe discriminative features, e.g. from medical images and other dense images.
\item The heatmaps produced by ResNet appear to be sparsest, followed by VGG then AlexNet. Input size and depth of networks may be the reasons.
\item A research into the role of negative values in the heatmaps may be necessary. If we continue with the interpretation that negative values correspond to negative contribution of prediction, some XAI methods such as DeepLift and DeepLiftShap may be completely incomprehensible.
\item There are phenomena where large continuous patches or areas are ignored and where red and blue signals are mixed in close proximity. More investigations may be needed, e.g. on how to find the best \textit{channel adjustment}.
\end{itemize}

\noindent In practice, it may not be easy to handle various noises produced by XAI methods. However, it is not impossible. Here we recommend that, in practical applications, every stakeholder should first agree on what constitutes an explanation. Our framework advocates the use of localization and distinguishing features as explanations. However, localization is sometimes sufficient as the explanation in computer vision problems, which is acceptable if agreed upon by all stakeholders. Otherwise, a major feature is to be selected as the main explanation. 

 We have provided an algorithm to produce synthetic data that we hope can be a baseline framework for testing XAI method, especially in the form of saliency maps or heatmaps. Some XAI methods appear to be more suitable for localization, while others are more responsive to the edges of features. In addition, mabCAM has been introduced as the framework compatible with the computation of multi-banded attribution heatmaps such as our synthetic data. The process is similar to semantic segmentation, a task which DNN has proven capable of performing.

Modifications required to boost the explainability power of existing XAI methods might differ across methods, making fair comparison a possibly difficult task. At least, for each application of XAI method, we should attempt to find a clear, consistent interpretation under the same context of study. For example, if negative values need to be treated with absolute values in some applications, at least an accompanying experiment is needed to show the effects and implications of performing such transformation. As of now, XAI still remains a challenging problem. However, it does exhibit good potential to improve the reliability of black-box models in the future.

\textit{Future directions}. Synthetic data with clear-cut shapes is useful as a yardstick to measure the quality of XAI methods in an objective manner. Our synthetic dataset allows readers to clearly see the pros and cons of using certain XAI methods. We have also included rich features in our dataset enough to simulate the variety found in real life dataset, including background noises. We hope that the practice of including magnitude-banded heatmaps can be expanded in the future, and it may be useful to craft explanation ground-truth labels with magnitude banding on real datasets. Further studies on the performance of mabCAM on other datasets will help improve the objectivity of heatmap as explanation framework as a whole.

\section*{Appendix}
The links to our appendix section is available in our github.

\begin{table}[h]
\begin{tabular}{c c}
\hline
AI & Artificial Intelligence  \\ 
AOPC& Area under Perturbation Curve\\ 
CAM & Class Activation Mapping  \\ 
CHR & Computational vs human relevance\\ 
CNN & Convolutional Neural Network \\ 
DNN & Deep Neural Network  \\ 
FC& Fully connected \\ 
FN& False Negative \\ 
FP& False Positive\\
GradCAM & Gradient CAM\\ 
ILSVRC & ImageNet Large Scale Visual Recognition Challenge \\ 
IoU&Intersection over Union \\
 LIME & Local Interpretable model-agnostic explanation \\
 LRP& Layerwise Relevance Propagation \\
 mab&magnitude-banded\\
 MoRF& most relevant first \\
 ROAR&RemOve and Retrain \\
 ROC&Receiver OPerating Characteristics\\
 SHAP&Shapley Additive exPlanation \\
 SPA & Spatial Product Attention \\
 TN&True Negative \\
 TP& True Positive \\
  XAI & eXplainable AI \\
\hline
\end{tabular}
\end{table}

\section*{Acknowledgment}

This research was supported by Alibaba Group Holding Limited, DAMO Academy, Health-AI division under Alibaba-NTU Talent Program, Alibaba JRI. The program is the collaboration between Alibaba and Nanyang Technological university, Singapore. This work was also supported by the RIE2020 AME Programmatic Fund, Singapore (No. A20G8b0102).

%
%

\bibliographystyle{unsrtnat}
\bibliography{xaib_TAI}

\newpage

\textbf{Training}. \textit{Regarding the dimensions}. We intend to use HxW=512x512 for all trainings, but it turned out AlexNet and VGG were not very efficient on the relatively small GPU (Tesla K40), hence we make them easier. 

Batch sizes are limited by our GPU we used for fine-tuning. The determination of \(r_f\) is simple: from our trial and error, if we found convergence harder to attain, we lower \(r_f\) so the model can be trained longer. 

\textbf{Hyperparameters and arbitrary settings}. As usual, hyperparameters can often be very arbitrarily chosen. However, we will stick to the following philosophy: given good models that predict accurately (accurate enough in our arbitrary opinion), then we ask, are their XAI heatmaps good? If we find that a good predictive model does not necessarily yield good heatmaps, then we can assert that XAI methods may perform poorly, depending on the settings. We wish to show that these existing XAI methods might still need improvements (consistent with our general conclusion). 

\textit{Information loss in channel adjustment}. During training, the colour is not adjusted away. Only during evaluation did we sum away the channel “contribution”. We do this based on the prevailing idea that positive/negative heatmap pixels correspond to positive/negative attributions. If readers consider our results very poor, then this paper achieves one of its objectives: it is partially intended to show readers that “standard” heatmaps may not perform well. We have therefore suggested that “a research into the role of negative values in the heatmaps may be necessary” (see conclusion).

\subsection*{Algorithms}

\parskip0em 
\textbf{Algorithm 1}, some details and descriptions:
\begin{enumerate}
\item $t$ is the thickness of cell border. 
\item Subscript ex stands for ``explanation", which will be the heatmap parts.
\item Thresholds $th_d, th_l=0.05$ are suitably chosen to create binary arrays.
\item $y_s$ is a multiplicative factor for modifying object's elliptical shape. 
\item Noise term. Noise added to \(d\) can be found in pipeline.objgen.cells (see code). Noise is given by normally distributed random array of size \(int(dn/noise_roughness)\) resized to the size of \(d\) where the default \(noise_roughness=0.1\).
\end{enumerate}
\parskip1em 

\parskip0em 
\textbf{Algorithm 2}, some details and descriptions:
\begin{enumerate}
\item $v_s$ is a multiplicative factor for stretching. 
\item $t_b,t_p$ are bar and pole thicknesses to form minus- and plus- shaped skeletons of the cells.
\end{enumerate}
\parskip1em 

\subsection*{More results and figures}

Heatmap figures shown in the main texts are not representative of all heatmaps predicted by any particular XAI method and any architecture. Some figures have slightly distorted shapes (see main appendix). Nevertheless, the pixel granularities of heatmaps generated by the same XAI methods are similar. All heatmaps above are obtained from correctly predicted samples

\setcounter{figure}{0}
\begin{figure}[htpb]
\includegraphics[width=0.5\textwidth,]{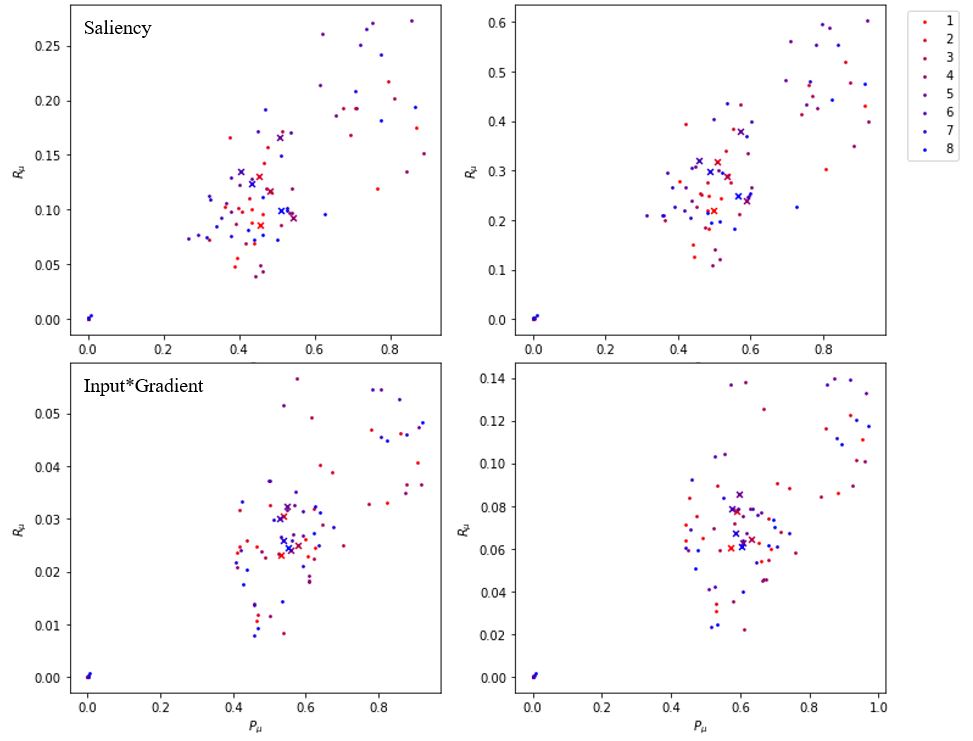}
\caption{Precision and recall values. Each colored dot is an averaged value of a given predicted class, each colored x mark the mean over all these dots given the same no. of additional convolutional layers. There is no significant observable trend. The number 1 to 8 refers to the number of additional convolutional layers added to the AlexNet.}
\label{fig:appdx1}
\end{figure}

\begin{figure}[htb!]
\centerline{\includegraphics[width=0.4\textwidth, trim={0 0 10cm 0},clip ]{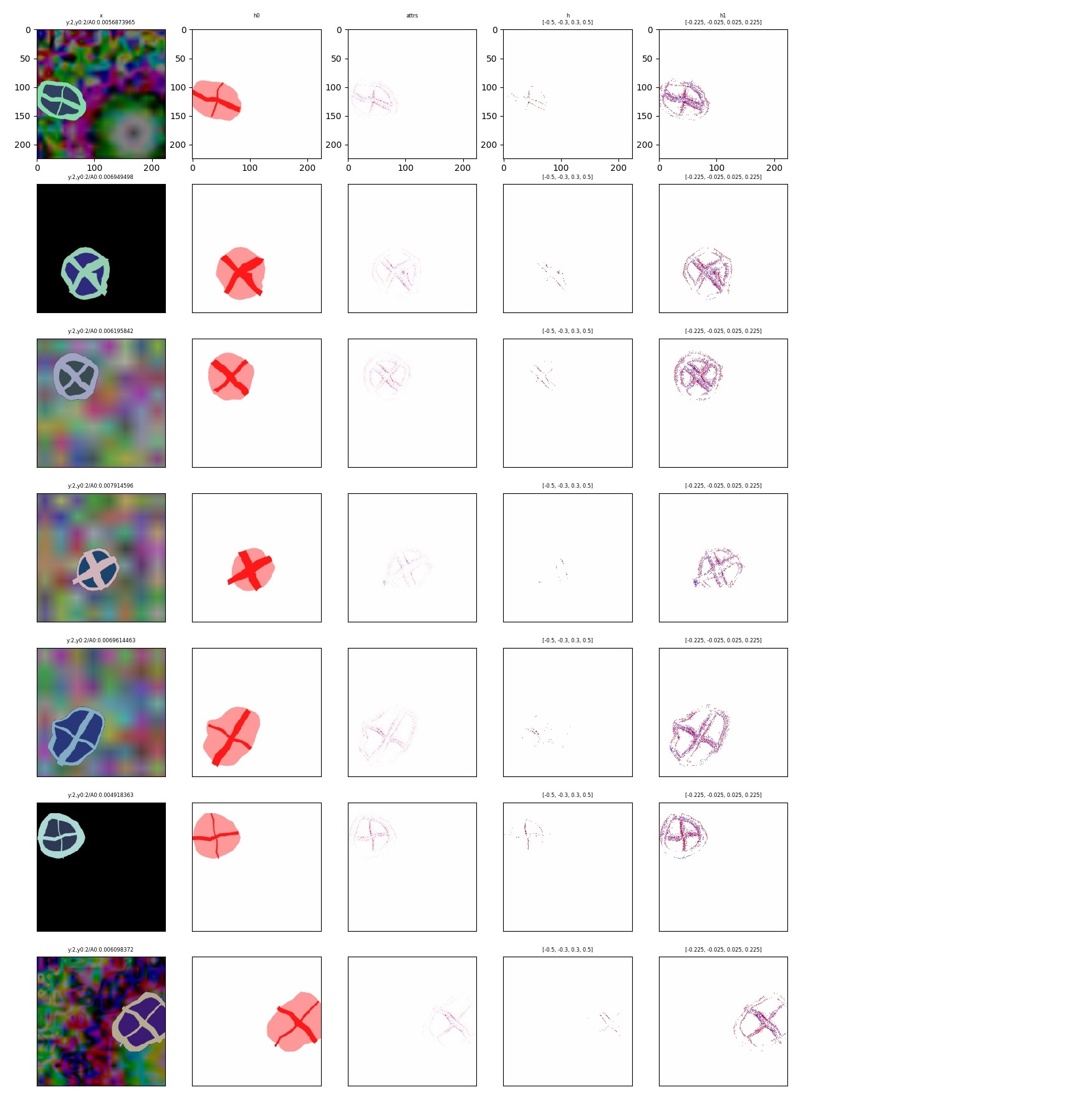}}
\caption{ResNet, DeepLiftShap. (This is fig. 17 in the main appendix)}
\end{figure}

\begin{figure}[htb!]
\centerline{\includegraphics[width=0.4\textwidth, trim={0 0 10cm 0}, clip]{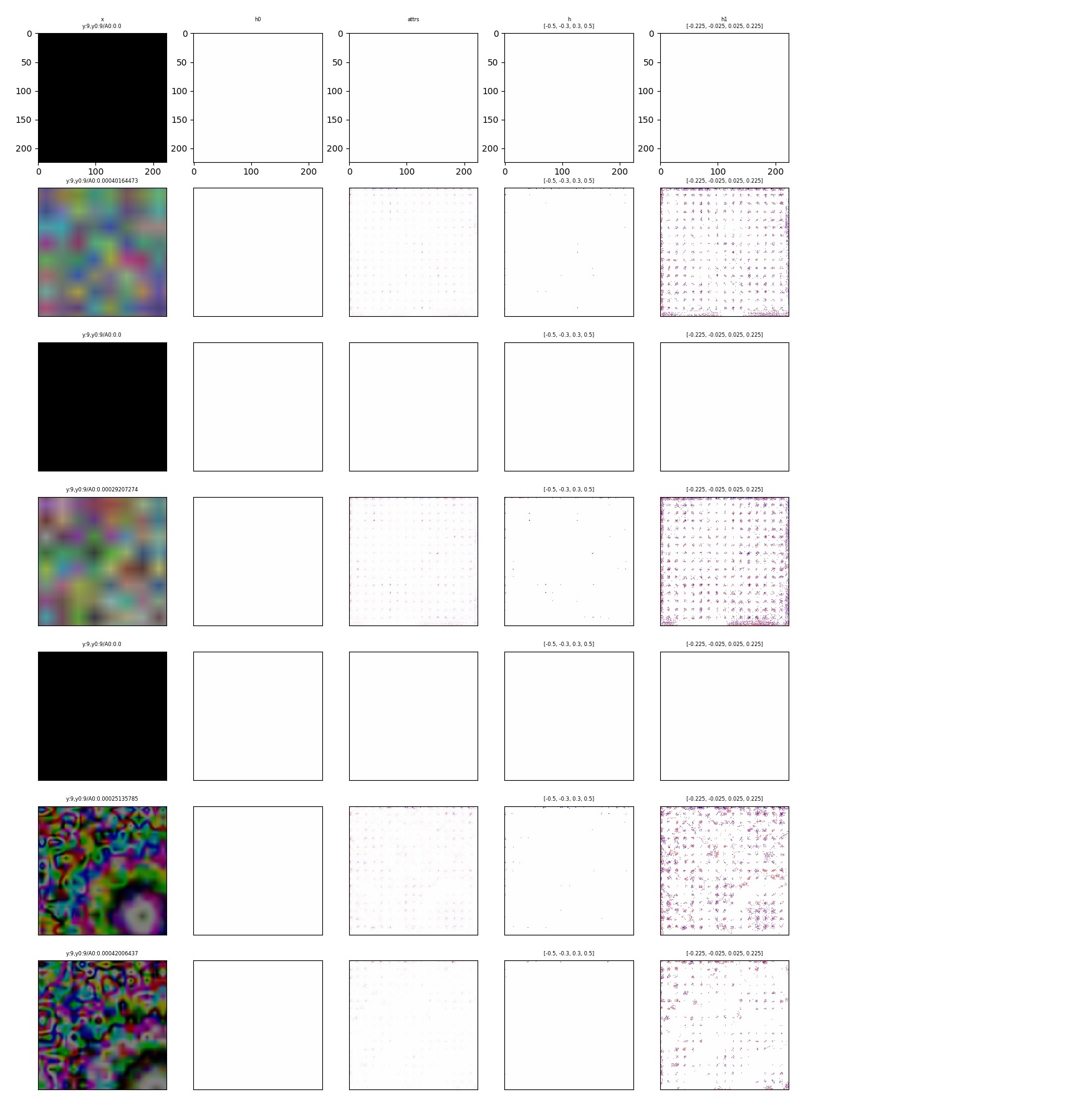}}
\caption{Resnet, Input*Gradient. (This is fig. 40 in the main appendix)}
\end{figure}
%

\subsection{mabCAM on three-class data}
\begin{figure*}[htpb]
\centerline{\includegraphics[width=2\columnwidth]{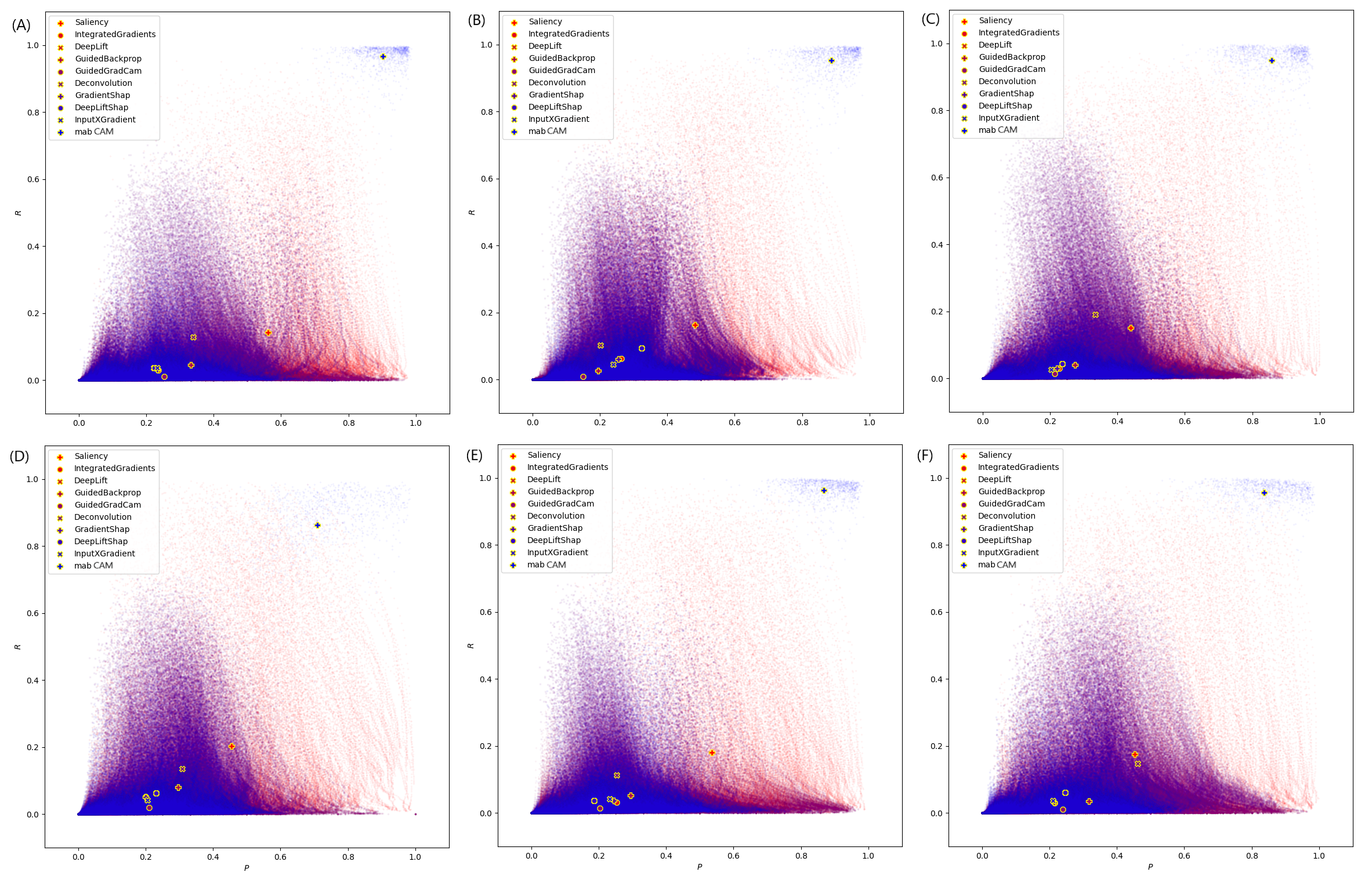}}
\caption{Recall versus precision scores of five-band stratified heatmaps compared to ground-truth for mabCAM. Similar to fig. \ref{fig:mabcam_main_result} in the main text. Zoomed in view on an electronic media is highly recommended. (A) \(\lambda_{mab}=5\), (B-F) \(\lambda_{mab}=10\)}
\label{fig:mabcam_other_results}
\end{figure*}

In the main text, mabCAM experiment is performed on ten-class dataset. Here, we show results for a similar experiment with \(\lambda_{mab}=5\) and five other trials with \(\lambda_{mab}=10\); all settings can be found in the github repository. In all results, validation accuracy of 0.975 is achieved and thus training is stopped early. Test accuracy is around 0.98. Most heatmap accuracy results are similar to the main text, as shown in fig. \ref{fig:mabcam_other_results}. In short, on average macCAM achieves recall and precision scores, typically above 0.8, while other XAI methods achieve much lower scores.

\end{document}